\documentclass{article}

\usepackage{arxiv}
\usepackage{latexsym}
\usepackage{graphicx}
\usepackage{multicol,multirow}
\usepackage{amsmath,amssymb,amsfonts}
\usepackage{mathrsfs}
\usepackage{amsthm}
\usepackage{rotating}
\usepackage{appendix}
\usepackage[sort&compress,numbers]{natbib}

\usepackage{ifpdf}
\usepackage[T1]{fontenc}
\usepackage{times}
\usepackage{sourcesanspro}
\usepackage{newtxmath}
\usepackage{textcomp}%
\usepackage{xcolor}%
\usepackage{hyperref}
\usepackage{lipsum}
\usepackage{calc}
\DeclareMathOperator{\sech}{sech}

\title{Long-term stability and generalization of observationally-constrained stochastic data-driven models for geophysical turbulence}
\author{Ashesh Chattopadhyay \\
Rice University\\
Houston, TX 77005 \\
\And
Jaideep Pathak \\
NVIDIA Corporation\\
Santa Clara, CA 95051 \\
\And
Ebrahm Nabizadeh \\
Rice University\\
Houston, TX 77005 \\
\And
Wahid Bhimji \\
Lawrence Berkeley\\ National Laboratory \\
Berkeley, CA 94720 \\
\AND
Pedram Hassanzadeh \\
Rice University \\
Houston, TX 77005 \\
}

\date{}

\begin{document}
\maketitle
\begin{abstract}
Recent years have seen a surge in interest in building deep learning-based fully data-driven models for weather prediction. Such deep learning models if trained on observations can mitigate certain biases in current  state-of-the-art weather models, some of which stem from inaccurate representation of subgrid-scale processes. However, these data-driven models, being over-parameterized, require a lot of training data which may not be available from reanalysis (observational data) products. Moreover, an accurate, noise-free, initial condition to start forecasting with a data-driven weather model is not available in realistic scenarios. Finally, deterministic data-driven forecasting models suffer from issues with long-term stability and unphysical climate drift, which makes these data-driven models unsuitable for computing climate statistics. Given these challenges, previous studies have tried to pre-train deep learning-based weather forecasting models on a large amount of imperfect long-term climate model simulations and then re-train them on available observational data. In this paper, we propose a convolutional variational autoencoder-based stochastic data-driven model that is pre-trained on an imperfect climate model simulation from a 2-layer quasi-geostrophic flow and re-trained, using transfer learning, on a small number of noisy observations from a perfect simulation. This re-trained model then performs stochastic forecasting with a noisy initial condition sampled from the perfect simulation. We show that our ensemble-based stochastic data-driven model outperforms a baseline deterministic encoder-decoder-based convolutional model in terms of short-term skills while remaining stable for long-term climate simulations yielding accurate climatology.

\end{abstract}

\section[Introduction]{Introduction}

A surge of interest in building deep learning-based data-driven models for complex systems such as chaotic dynamical systems \cite{pathak2018model,chattopadhyay2020data}, fully turbulent flow \cite{chattopadhyay2020deep}, and weather and climate models \cite{weyn2019can,weyn2020improving,weyn2021sub,rasp_2020_resnet,chattopadhyay2021towards} has been seen in the recent past. This interest in data-driven modeling stems from the hope that if these data-driven models are trained on observational data, (i) they would not suffer from some of the biases of physics-based numerical climate and weather models \citep{balaji2021climbing,schultz2021can}, e.g., due to ad-hoc parameterizations, (ii) they can be used to generate large ensemble forecasts for data assimilation \cite{chattopadhyay2021towards}, and (iii) they might seamlessly be integrated to perform climate simulations \cite{watson2021machine} that would allow for generating large synthetic datasets to study extreme events \citep{chattopadhyay2019analog}. 

Despite the promise of data-driven weather prediction  (DDWP) models, most DDWP models cannot compete with state-of-the-art numerical weather prediction (NWP) models \citep{scher2019weather,rasp_2020_resnet,weyn2019can,weyn2020improving,chattopadhyay2021towards} although more recently FourCastNet \cite{pathak2022fourcastnet} has shown promise of being quite competitive with the best NWP models. There have also been some previous work, where pre-training models on climate simulations have also resulted in improved short-term forecasts on reanalysis datasets \cite{rasp_2020_resnet}. Moreover, these DDWP models, while comparable with NWP in terms of short-term forecasts, are unstable when integrated for longer time scales. These instabilities either appear in the form of blow-ups or unphysical climate drifts \citep{scher2019weather,chattopadhyay2020deep}. 


In this paper, we consider the following realistic problem set-up, wherein, we have long-term climate simulations from an imperfect 2-layer quasi-geostrophic (QG) flow (we will call it "imperfect system" from here on) at our disposal to pre-train a DDWP. Following that, we have a few noisy observations from a perfect 2-layer QG flow ("perfect system" from here on) on which we are able to fine-tune our DDWP. This observationally-constrained DDWP model is then expected to generalize to the perfect system with a noisy initial condition (obtained from the available observations) and perform stochastic short-term forecasting. It is also expected to be seamlessly time-integrated to obtain long-term climatology. We show that in this regard, a stochastic forecasting approach with a generative model such as a convolutional variational autoencoder (VAE) outperforms a deterministic encoder-decoder-based convolutional architecture in terms of short-term forecasts and remains stable without unphysical climate drift for long-term integration. A deterministic encoder-decoder would fail to do so and shows unphysical climate drift followed by numerical blow-up.

\section{Methods}

\subsection{Imperfect and Perfect Systems}

In this paper, we consider 2-layer QG flow as the geophysical system on which we intend to perform both short-term forecasting and compute long-term statistics. 

The dimensionless dynamical equations of the 2-layer QG flow are derived following Lutsko et al.~\cite{lutsko2015applying}, and Nabizadeh et al.~\cite{nabizadeh2019size}. The system consists of two constant density layers with a $\beta$-plane approximation in which the meridional temperature gradient is relaxed towards an equilibrium profile. The equation of the system is described as:

\begin{align}
\label{eq_QG}
\begin{split}
    \frac{\partial q_j}{\partial t}+J(\psi_j,q_j)=-\frac{1}{\tau_d}(-1)^{j}\left(\psi_1-\psi_2-\psi_R\right) \\
    -\frac{1}{\tau_f}\delta_{k2}\nabla^2\psi_j - \nu \nabla^{8}q_j.
       \end{split}
\end{align}
Here, $q$ is potential vorticity and is expressed as
\begin{align}
\label{q_def}
q_j=\nabla^2\psi_j +(-1)^{j}\left(\psi_1-\psi_2\right)+\beta y,
\end{align}
where $\psi_j$ is the stream function of the system.
In Eqs.~(\ref{eq_QG}) and~(\ref{q_def}), $j$ denotes the upper ($j=1$) and lower ($j=2$) layers.~$\tau_d$ is the Newtonian relaxation time scale while $\tau_f$ is the Rayleigh friction time scale, which only acts on the lower layers represented by the Kronecker $\delta$ function ($\delta_{k2}$).~$J$ denotes the jacobian, $\beta$ is the $y-$gradient of the coriolis parameter, and $\nu$ denotes the hyperdiffusion coefficient. We have induced a baroclinically unstable jet at the center of a horizontally (zonally) periodic channel by setting $\psi_1-\psi_2$ to be equal to a hyperbolic secant centered at $y=0$. When eddy fluxes are absent, $\psi_2$ is identically zero, making zonal velocity in the upper layer $u(y)=-\frac{\partial \psi_1}{\partial y}=-\frac{\partial \psi_R}{\partial y}$ where we set
\begin{align}
\label{init_eq}
-\frac{\partial \psi_R}{\partial y}=\sech^2\left(\frac{y}{\sigma}\right),
\end{align}
$\sigma$ being the width of the jet. Parameters of the model are set following the previous studies~\cite{lutsko2015applying,nabizadeh2019size}, in which $\beta= 0.19$, $\sigma=3.5$, $\tau_f=15$, and $\tau_d=100$. 

To non-dimensionalize the equations, we have used the maximum strength of the equilibrium velocity profile as the velocity scale ($U$) and the deformation radius ($L$) for the length scale. The system’s time scale ($L/U$) is referred to as the “advective time scale” ($\tau_{adv}$).

The spatial discretization is spectral in both $x$ and $y$ where we have retained $96$ and $192$ Fourier modes, respectively. The length and width of the domain is equal to $46$ and $68$ respectively after non-dimensionalizing the numbers. The spurious waves on the northern and southern boundaries are damped by applying sponge layers. Note that the domain is wide enough that the sponges do not affect the dynamics. Here $5 \tau_{adv} \approx$ $1$ Earth day $\approx 200 \Delta t_n$, where $\Delta t_n=0.025$ is the time step of the leapfrog time integrator used in the numerical scheme. 

\begin{figure*}[t]
\includegraphics[width=\textwidth]{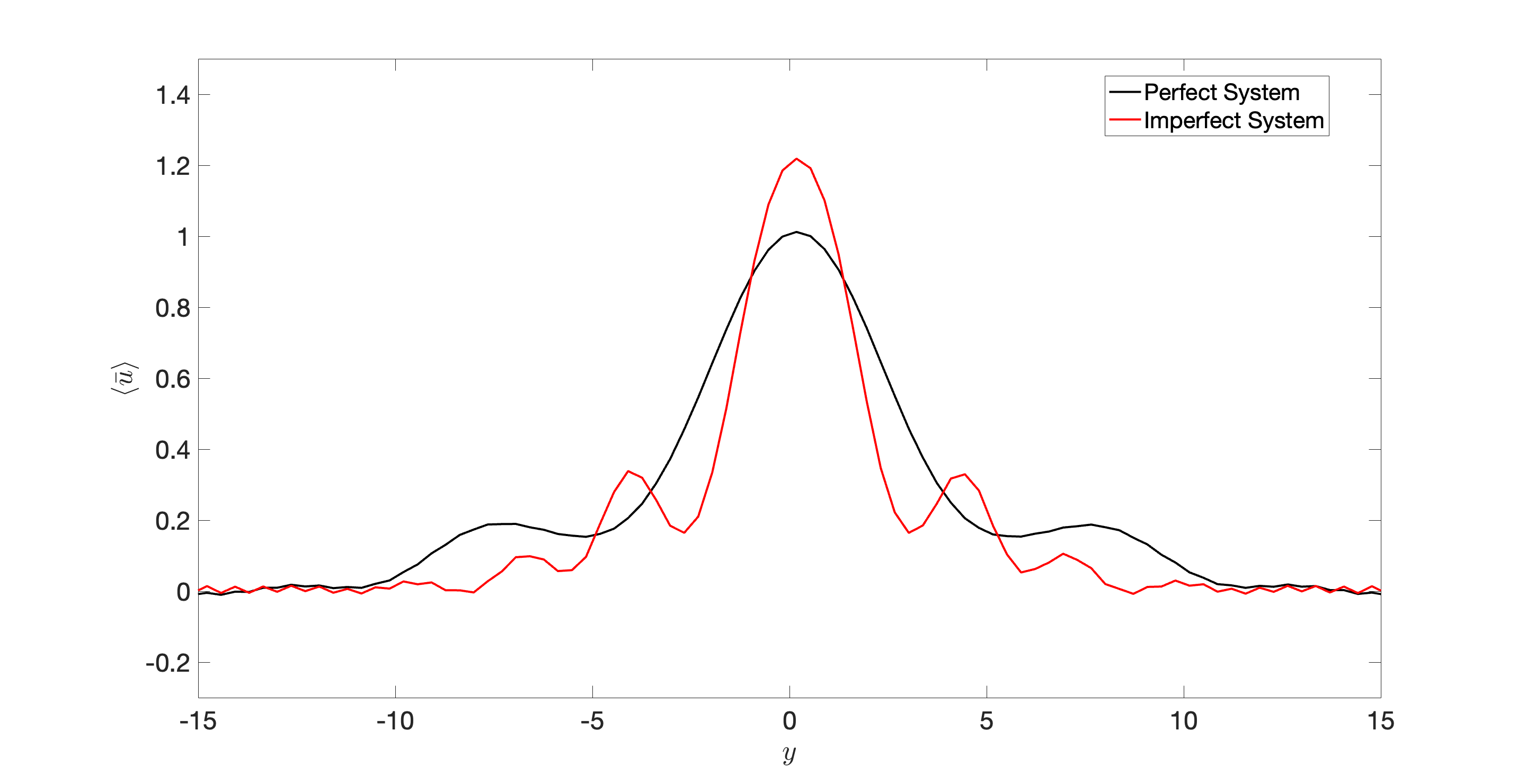}
\caption{Time-averaged zonal-mean velocity (over $20000$ days), $\left<\bar{u}\right>$, of the imperfect and perfect system. The difference between $\left<\bar{u}\right>$ of the perfect and imperfect systems indicates a challenge for DDWP models to seamlessly generalize from one system to the other}
\label{2-system}
\end{figure*}

The imperfect system in this paper has an increased value of $\beta$, given by $\beta^* = 3\beta$ and a decreased size of the jet, given by $\sigma*=\frac{4}{5}\sigma$. We assume that we have long-term climate simulations of the imperfect system while we have only a few noisy observations of the perfect system. The difference in the long-term averaged zonal-mean velocity in the upper layer of the perfect and imperfect systems is shown in Fig.~\ref{2-system}. It is to be noted that the motivation behind having an imperfect and perfect system is the fact that we assume that the \textit{imperfect system} represents a climate model with biases from which we have long simulations at our disposal, while the \textit{perfect system} is analogous to actual observations from our atmosphere, where the sample size is small and contaminated with measurement noise.

\subsection{VAE and Transfer Learning Frameworks}
Here, we propose a generative modeling approach to perform both short- and long-term forecasting for the perfect QG system after being trained on the imperfect QG system. We assume that we have a long-term climate simulation from the imperfect system, with states $x^m(k\Delta t)$, where $k\in{1,2,\cdots K}$ and $x^m \in \mathcal{R}^{2\times 96\times 192}$. A convolutional VAE is trained on the states of this imperfect system to predict an ensemble of states with a probability density function, $p\left(x^m\left(t+\Delta t\right)\right)$, from $x^m(t)$. Here, $\Delta t = 40 \Delta t_n$, which is also the sampling interval for training the model. Extensive trial and error-based search has been performed for hyperparameter optimization for the VAE. The details on the VAE architecture is given in Table~\ref{tab:VAE}.

\begin{figure*}
\includegraphics[width=\textwidth]{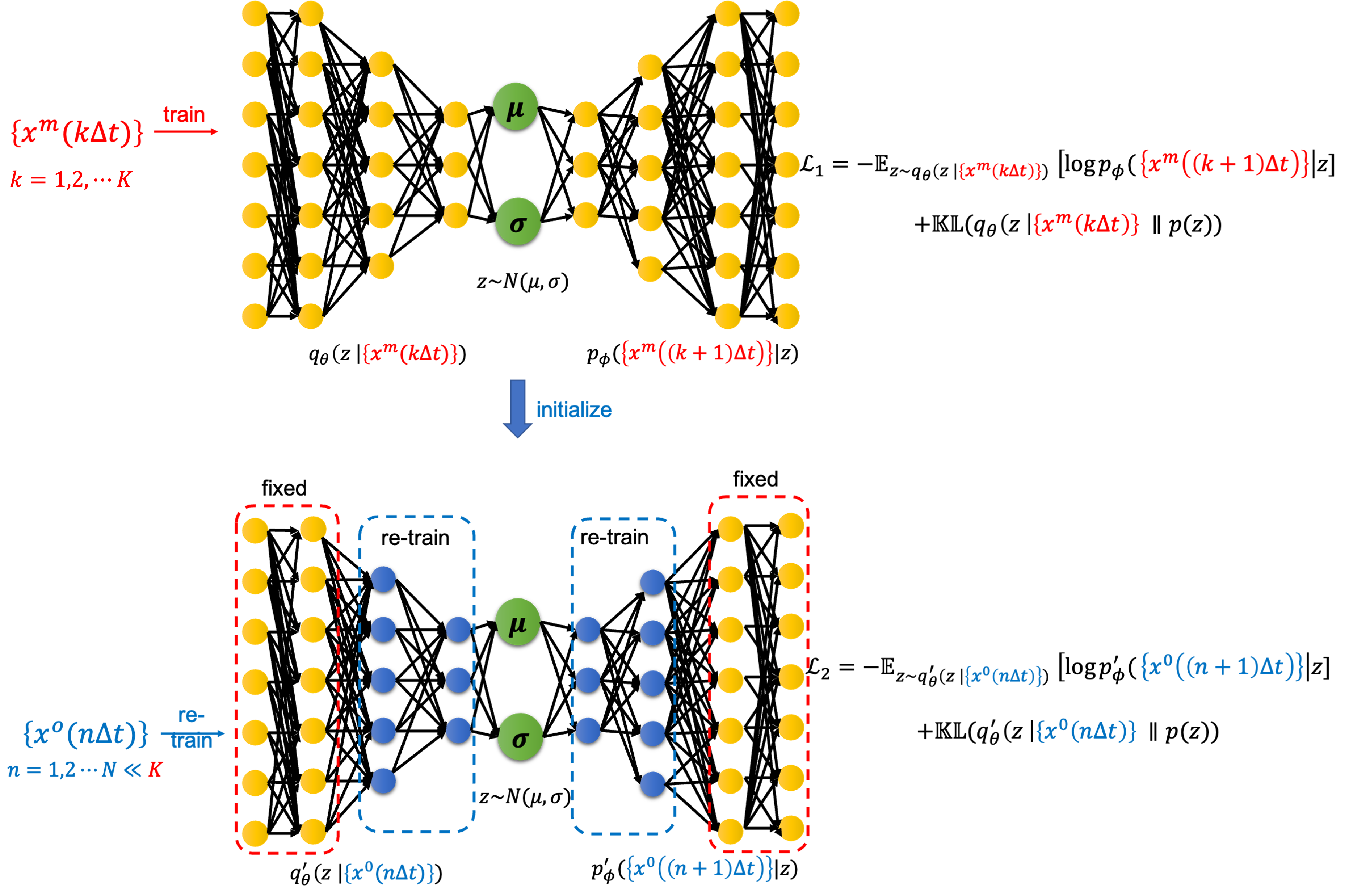}
\caption{A schematic of the transfer learning framework with a VAE pre-trained on imperfect simulations and transfer learned on noisy observations from the perfect system. Here, $x^m\left(k\Delta t\right)$ are states obtained from the imperfect system and $x^o\left(n\Delta t\right)$ are noisy observations from the perfect system. Here, $\Delta t = 40 \Delta t_n$. Note that, our proposed VAE is convolutional and the schematic is just representative. Details about the architecture are given in Table~\ref{tab:VAE}}
\label{framework}
\end{figure*}

This pre-trained VAE would then undergo transfer learning on a small number of noisy observations of the states of the perfect system $x^o(n\Delta t)$ where $n\in{1,2,\cdots N}$ and $N \ll K$. The VAE would then stochastically forecast an ensemble of states of the perfect system with a noisy initial condition from the perfect system. The noise in the initial condition is sampled from a Gaussian normal distribution with $0$ mean and standard deviation being $\eta \sigma_Z$, where $\sigma_Z$ is the standard deviation of $\psi_k$ and $\eta$ is a fraction that determines the amplitude of the noise vector. A schematic of this framework is shown in Fig.~\ref{framework}. In this work, we consider the stream function $\psi_k$ ($k=1,2$) to be the states of the system on which we would train our VAE. 

The VAE is trained on $9$ independent ensembles of the imperfect system with $1400$ consecutive days of climate simulation each with a sampling interval of $\Delta t$. Hence, the training size for the VAE is about $12600$ days of data. For transfer learning, we assume that only $10\%$ of the number of training samples is available as noisy observations from the perfect system. 

\begin{table}[]
    \centering
    \begin{tabular}{|l||l|p{0.1\textwidth}|}
    \hline
          \textbf{Number}& \textbf{Layer} & \textbf{Number of Filters} \\
         \hline
         1 & $5\times 5$ 2D Convolution & 32 \\
         \hline
         \hline
         2 & $2\times 2$ Max Pooling & --\\
         \hline
         3 & $5\times 5$ 2D Convolution & 32 \\
         \hline
         \hline
         4 & $2\times 2$ Max Pooling & -- \\
         \hline
         5 & $5\times 5$ 2D Convolution & 32 \\
         \hline
          6 & $2\times 2$ Max Pooling & -- \\
         \hline
         7 & flatten & --  \\
         \hline
         8 & 128 neurons for mean and standard deviation & -- \\
         \hline
         9 & Dense layer & --\\
         \hline
         10 & $5\times 5$ 2D Convolution & 32\\
          \hline
          11 & Up-sampling & --\\
          \hline
         12 & $5\times 5$ 2D Convolution & 32 \\
          \hline
         13 &  Up-sampling & --\\
          \hline
        14 & $5\times 5$ 2D Convolution & 32 \\
          \hline
         15 &  Up-sampling & --\\
         \hline
                 15 & $5\times 5$ 2D Convolution & 32 \\
          \hline
         16 &  Up-sampling & --\\
         \hline
         17 & $5\times 5$ 2D Convolution & 2 \\
         \hline
    \end{tabular}
    \caption{Number of layers and filters in the convolutional VAE architecture used in this paper. The baseline convolutional encoder-decoder model would have the exact same architecture and the same latent space size as that of the VAE. }
    \label{tab:VAE}
\end{table}

\section{Results}

\subsection{Short-term Stochastic Forecasting on the Imperfect System}
\begin{figure*}[t]
\includegraphics[width=\textwidth]{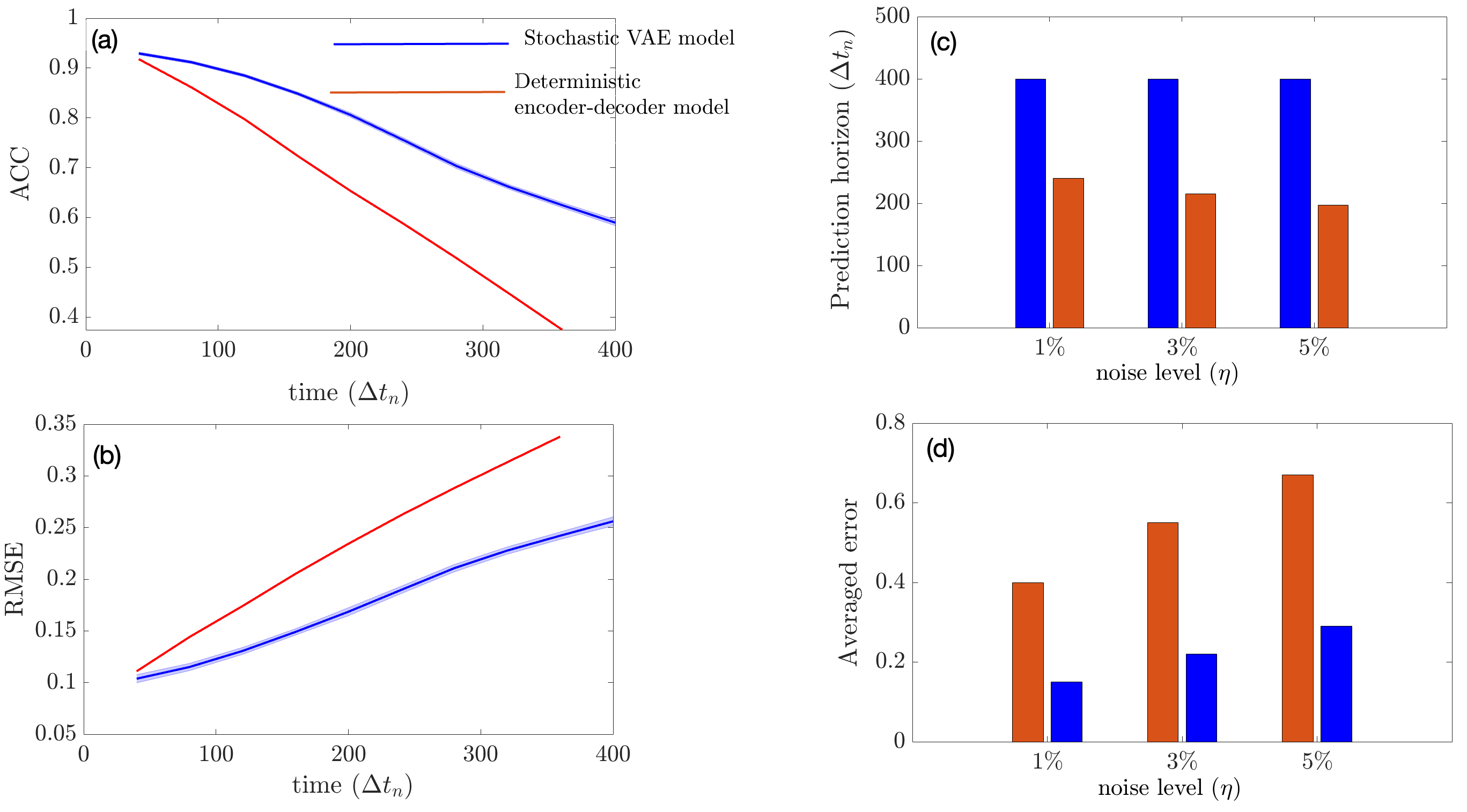}
\caption{Performance of the stochastic convolutional VAE as compared to baseline convolutional encoder-decoder-based model when trained on the imperfect system and predicts from a noisy initial condition sampled from the imperfect system.
(a) Anomaly correlation coefficient (ACC) between predicted $\psi_1$ and true $\psi_1$ \citep{murphy1989skill} from the imperfect system with $\eta = 5\%$ (noise level) for the initial condition. (b) Same as (a) but for RMSE. (c) Prediction horizon (number of $\Delta t_n$ until ACC $\leq 0.60$) for different noise levels added to the initial condition. (d) Averaged error over $2$ days for different noise levels of initial condition. The shading shows the standard deviation across $100$ ensembles generated by the VAE model during inference}
\label{forecast_imperfect}
\end{figure*}
In this section, we show how well the stochastic VAE performs in terms of short-term forecasting on the imperfect system when the initial condition is sampled from the imperfect system. Our baseline model is a convolutional encoder-decoder-based deterministic model that has the exact same architecture as that of the convolutional VAE with the same size of the latent space. During inference, the stochastic VAE generates $100$ ensembles at every $\Delta t$ and the mean of these ensembles is fed back into the VAE autoregressively to predict future time steps. We see from Fig.~\ref{forecast_imperfect} that the stochastic VAE model outperforms the deterministic model. Our experiments suggest that increasing the noise level of the initial condition does not affect the prediction horizon of VAE if we consider ACC $= 0.60$ as the limit of prediction (although it is an ad-hoc choice) as shown in Fig.~\ref{forecast_imperfect}(c). This maybe attributed to the reduction of initial condition error (due to noise) with ensembling. The averaged RMSE error over $2$ days grows faster with an increase in the noise level in the deterministic model as compared to the stochastic VAE model as shown in Fig.~\ref{forecast_imperfect}(d).

\subsection{Short-term Stochastic Forecasting on the Perfect System}
\begin{figure*}
\includegraphics[width=\textwidth]{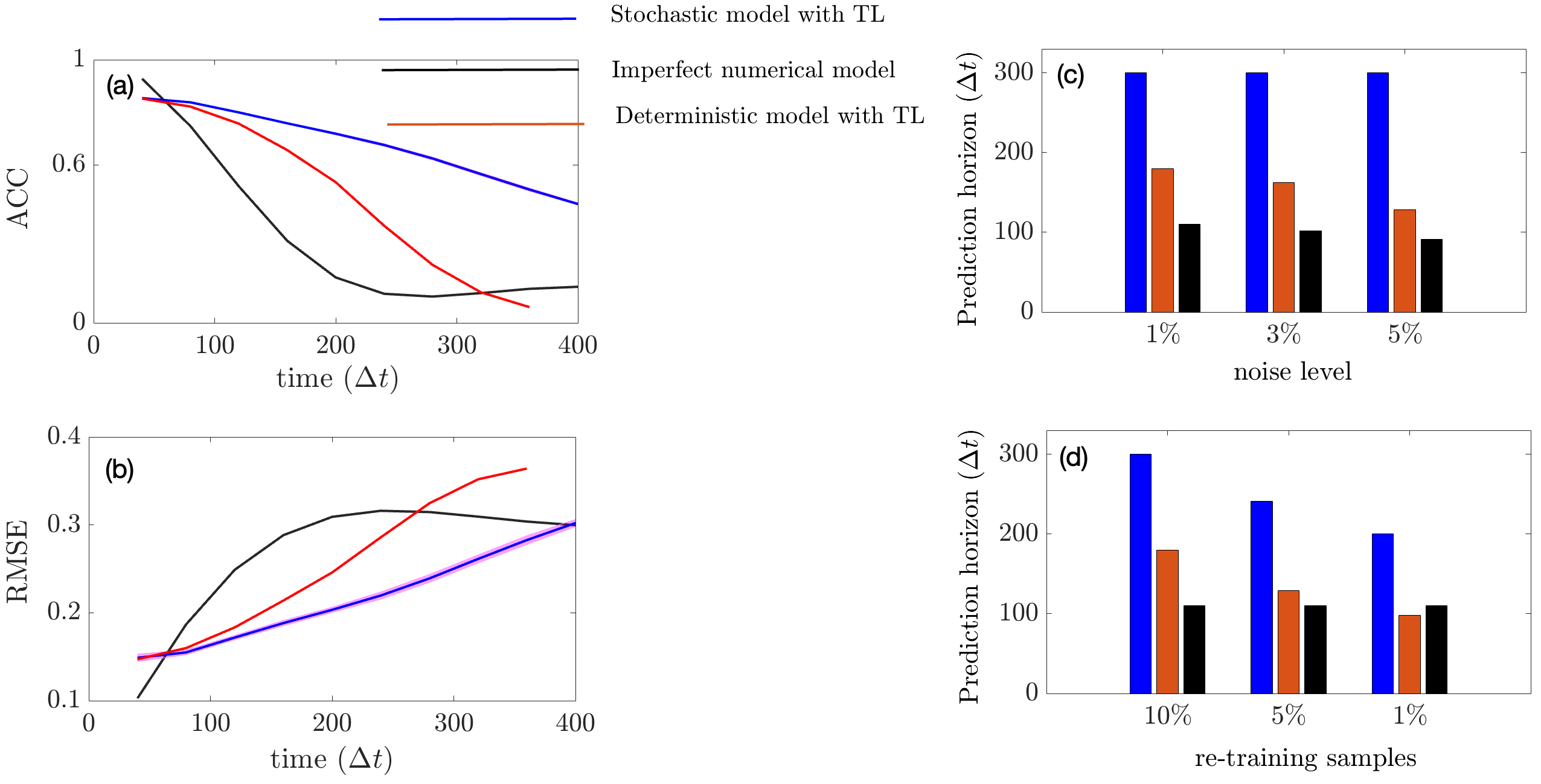}
\caption{Performance of the stochastic convolutional VAE compared to that of the baseline convolutional encoder-decoder-based model and the imperfect numerical model when trained on the imperfect system, re-trained on noisy observations from the perfect system, and initialized with a noisy initial condition sampled from the perfect system.
(a) Anomaly correlation coefficient (ACC) between predicted $\psi_1$ and true $\psi_1$ from the perfect system with $\eta = 5\%$ (noise level) for the initial condition. (b) Same as (a) but for RMSE. (c) Prediction horizon (number of $\Delta t_n$ until ACC $\leq 0.60$) for different noise levels added to initial condition. (d) Prediction horizon of the models with different sample sizes of noisy observation data (as percentages of original training sample size) from perfect system. Note that the numerical model is not trained on data and hence would show no effect.   }
\label{forecasts_perfect}
\end{figure*}

In this section, we show how well the pre-trained VAE (as compared to the baseline encoder-decoder and imperfect numerical model) on the imperfect climate simulations performs on the perfect system when constrained by noisy observations from the perfect system on which it is re-trained. Only two layers before and after the bottleneck layer are re-trained for both the VAE and the baseline model. Figure~\ref{forecasts_perfect} (a) and (b) show that for short-term forecasting, VAE outperforms both the baseline deterministic encoder-decoder model as well as the imperfect numerical model. Figure~\ref{forecasts_perfect}(c) shows a similar behaviour for VAE with increase in noise level in the initial condition as Fig.~\ref{forecast_imperfect}(c). Figure~\ref{forecasts_perfect}(d) shows the effect of the number of re-training samples used as noisy observations for transfer learning on the prediction horizon. While the numerical model is unaffected by the impact of noisy observations (since it does not undergo any learning), an increase in noise level to the initial condition makes it more susceptible to error as compared to the data-driven models which are inherently more robust to noise (due to re-training on noisy observations). In all cases, the VAE outperforms both the deterministic encoder-decoder model and the imperfect numerical model for short-term forecasts.       

\subsection{Long-term Climate Statistics}
\begin{figure*}[ht]
\includegraphics[width=\textwidth]{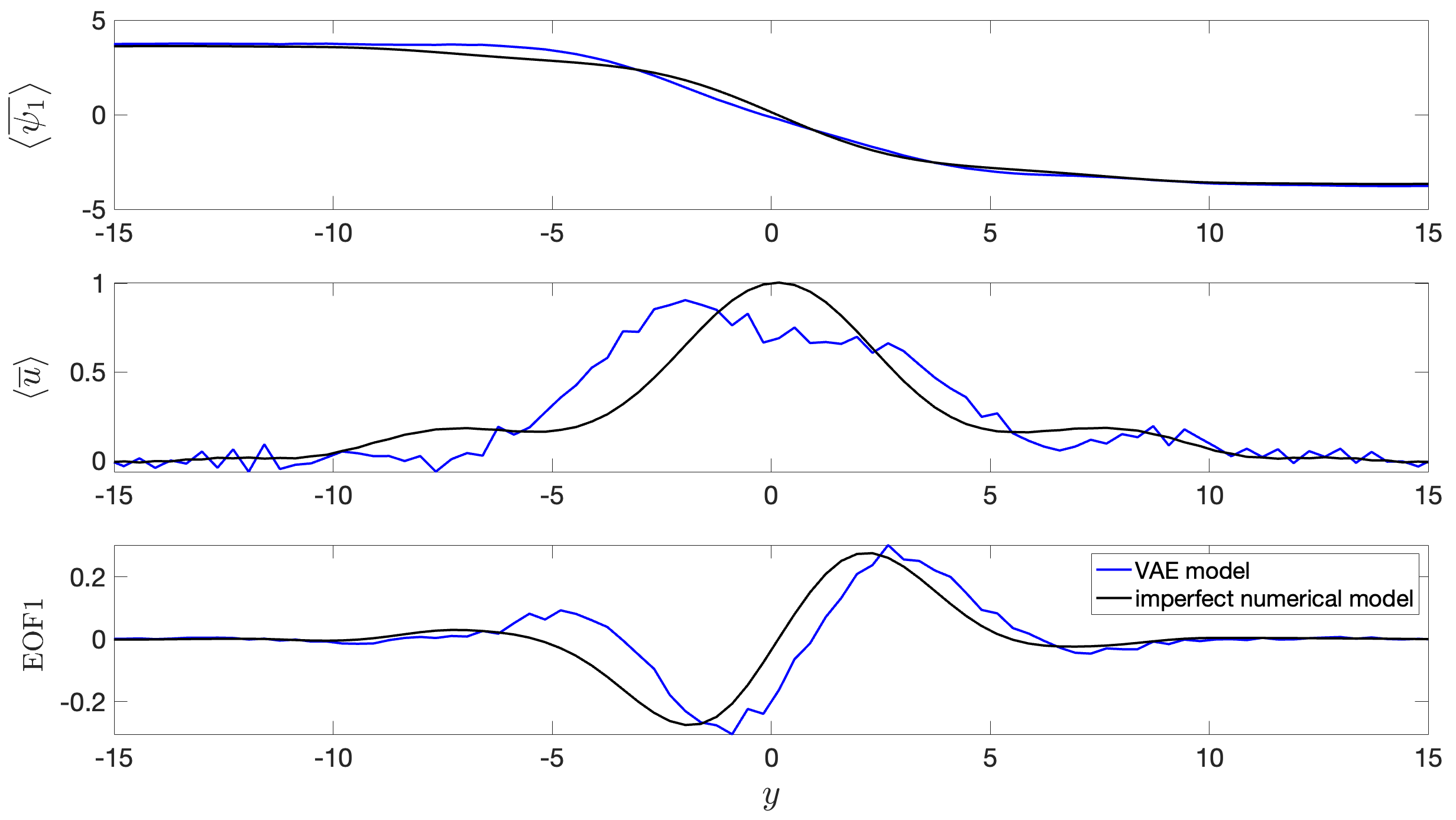}
\caption{Long-term climatology obtained from integrating the models for $20000$ days. $\left<\bar{\psi}_1\right>$ is the time mean zonally-averaged $\psi_1$, $\left<\bar{u}\right>$ is the time mean zonally averaged upper-layer velocity, and EOF1, is the first EOF (empiricial orthogonal function) of $\left<\bar{u}\right>$. The VAE shows non-drifting physical climate. Baseline encoder-decoder model is not shown since it becomes unstable within $100$ days of seamless integration. }
\label{long_term_forecasts}
\end{figure*}

In this section, we show the comparison between long-term climatology obtained from seamlessly integrating the VAE for 20000 days and the true climatology of the imperfect system. It must be noted that the deterministic encoder-decoder model is not stable and would show unphysical drifts in the climate similar to other deterministic data-driven models within a few days of seamless integration \citep{chattopadhyay2020deep}. Figure.~\ref{long_term_forecasts} shows stable and non-drifting physical climate obtained from the VAE whose mean of $\psi_j$, $u_j$, and EOF1 of $u_1$ closely match those of the true system. 

\section{Conclusion}
In this paper, we have developed a convolutional VAE for stochastic short-term forecasting and long-term climatology of the stream function of a 2-layer QG flow. The VAE is trained on climate simulations from an imperfect QG system before being fine-tuned on noisy observations from a perfect QG system on which it performs both short- and long-term forecasting. The VAE outperforms both the baseline convolutional encoder-decoder model as well as the imperfect numerical model. 

One of the main advantages of using the stochastic VAE with multiple ensemble members is the reduction of the effect of noise in the initial condition for short-term forecasting. Moreover, the VAE remains stable through a seamless $20000$ days integration yielding a physical climatology that matches the numerical model. Deterministic data-driven models do not remain stable for long-term integration and hence a stochastic generative modeling approach may be a potential candidate when developing data-driven weather forecasting models that can seamlessly be integrated to yield long-term climate simulations. 

Despite the improvement in performance in short-term skills as well as long-term stability, one of the caveats of the VAE is that it is less interpretable as compared to a deterministic model. Moreover, the VAE is more difficult to scale as compared to a deterministic model, especially for high-dimensional systems. This is because one would need to evolve a large number of ensembles for high-dimensional systems, and the exact relationship between short-term skills and long-term stability with the number of ensembles is not well known. Finally, while the application of VAE has resulted in long-term stability, the underlying causal mechanism by which instability is exhibited in a determinisitc data-driven model is still largely unknown. Future work needs to be undertaken to understand the cause of this instability such that determinisitc  models that may be more scalable than VAE can also be used for data-driven weather and climate prediction.   

\paragraph{Acknowledgments}
This work was started as an internship project by A.C. in Lawrence Berkeley National Laboratory in the Summer of 2021 under J.P. and W.B. It was then continued under P.H. as a part of A.C.'s PhD at Rice University. This research used resources of NERSC, a U.S. Department of Energy Office of Science User Facility operated under Contract No. DE-AC02-05CH11231.  A.C., E.N., and P.H. were supported by ONR grant N00014-20-1-2722, NASA grant 80NSSC17K0266, and NSF CSSI grant OAC-2005123. A.C. also thanks the Rice University Ken Kennedy
Institute for a BP HPC Graduate Fellowship. We are grateful to Kamyar Azizzadenesheli, Mustafa Mustafa, and Sanjeev Raja for providing invaluable advice for this work.

\paragraph{Competing Interests}
None

\paragraph{Data Availability Statement}
The codes for this work would be available to the public upon publication in this Github link: \url{https://github.com/ashesh6810/Stochastic-VAE-for-Digital-Twins.git}.

\paragraph{Ethical Standards}
The research meets all ethical guidelines, including adherence to the legal requirements of the study country.

\bibliographystyle{unsrtnat}
\bibliography{sample_arxiv}

\end{document}